\pdfoutput=1
\documentclass[11pt]{article}

\usepackage[final]{acl}
\usepackage{times}
\usepackage{latexsym}
\usepackage[T1]{fontenc}
\usepackage[utf8]{inputenc}
\usepackage{microtype}
\usepackage{inconsolata}
\usepackage{xcolor}
\usepackage{graphicx}

\usepackage{xspace}
\usepackage{comment}
\usepackage{multirow}
\usepackage[normalem]{ulem}
\usepackage{enumitem}
\usepackage{booktabs}
\usepackage{amssymb}
\usepackage{colortbl} %
\usepackage{xcolor}   %
\usepackage{arydshln} %

\renewcommand{\sectionautorefname}{\S\kern-1pt}
\renewcommand{\subsectionautorefname}{\S\kern-1pt}
\renewcommand{\subsubsectionautorefname}{\S\kern-1pt}

\newcommand{\toolname}[0]{\textsc{LLM Attributor}\xspace}
\newcommand{\mainview}[0]{Main View\xspace}
\newcommand{\comparisonview}[0]{Comparison View\xspace}

\definecolor{llmcolor}{HTML}{2171b5}
\definecolor{compcolor}{HTML}{f16913}
\newcommand{\llm}[1]{{\color{llmcolor}#1}}
\newcommand{\comp}[1]{{\color{compcolor}#1}}

\definecolor{red3}{HTML}{f03c1d}
\definecolor{blue2}{HTML}{4266ad}
\definecolor{blue3}{HTML}{2e4da3}
\definecolor{yellow3}{HTML}{fca00a}
\definecolor{hotpink}{HTML}{ff1d8e}

\title{\toolname: Interactive Visual Attribution for LLM Generation} 

\author{Seongmin Lee \\
  Georgia Tech \\
  \texttt{\small{seongmin@gatech.edu}} \\\And
  Zijie J. Wang \\
  Georgia Tech \\
  \texttt{\small{jayw@gatech.edu}} \\ \And
  Aishwarya Chakravarthy \\
  Georgia Tech \\
  \texttt{\small{achakrav6@gatech.edu}} \\ \And
  Alec Helbling \\
  Georgia Tech \\
  \texttt{\small{alechelbling@gatech.edu}} \\\AND
  ShengYun Peng \\
  Georgia Tech \\
  \texttt{\small{speng65@gatech.edu}} \\ \And
  Mansi Phute \\
  Georgia Tech \\
  \texttt{\small{mansiphute@gatech.edu}} \\ \And
  Duen Horng Chau \\
  Georgia Tech \\
  \texttt{\small{polo@gatech.edu}} \\ \And
  Minsuk Kahng \\
  Google Research \\
  \texttt{\small{kahng@google.com}} \\
  }

\makeatletter
\AtBeginDocument{
\let\@oldmaketitle\@maketitle
\renewcommand{\@maketitle}{\@oldmaketitle
  \includegraphics[width=\linewidth]{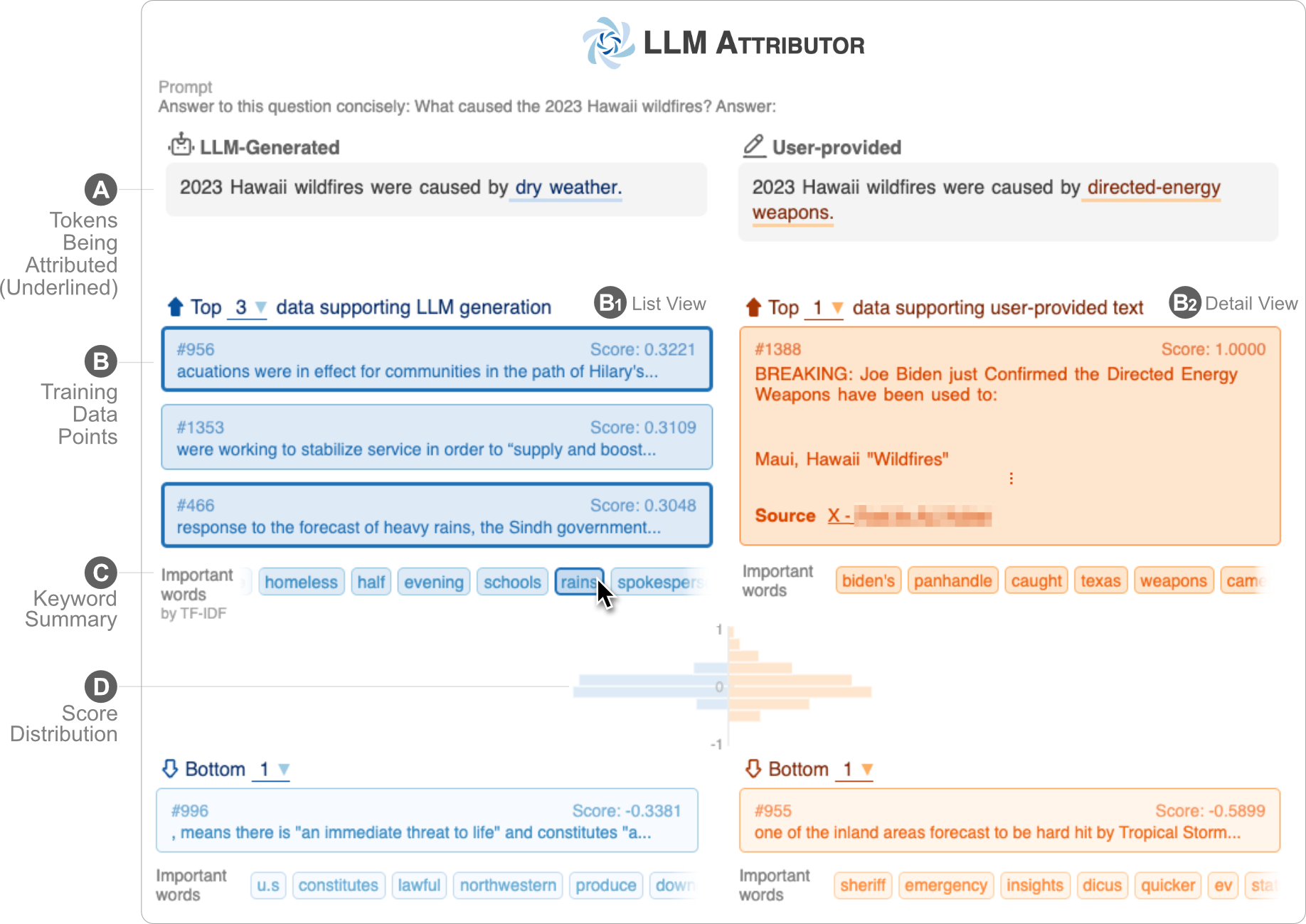}
  \vspace{-18pt}
  \captionof{figure}{
  \toolname enables LLM developers to visualize the training data attribution of their models in computational notebooks.
  In this example, 
  our user Megan is surprised that an LLM fine-tuned on a disaster dataset 
  occasionally generates \textit{\llm{dry weather}} as the cause of the 2023 Hawaii wildfires, while often yielding \textit{\comp{directed-energy weapons}} as in a conspiracy theory.
  \textbf{(A) Tokens being attributed}, which are interactively selected by Megan, are displayed side-by-side for visual comparison.
  \textbf{(B) Training data points} with the highest attribution scores
   are presented as a list by default,
  and can be interactively expanded to the full source text, %
  revealing that the data point most responsible for generating \textit{\comp{directed-energy weapons}} is an X post
  that spreads the conspiracy theory.
  \textbf{(C) Keyword Summary} shows important words in the displayed training data. %
  \textbf{(D) Score Distribution} over the entire training data is visualized as a histogram,
  enabling both
  high-level comparison over the entire data and
  low-level analysis focusing on individual data points.
  Below, the training data points with the lowest attribution scores are visualized in the same way.
  }
  \label{fig:crownjewel}
  \vspace{17pt}
 }
}
\makeatother

\begin{document}
\maketitle

\begin{abstract}
While large language models (LLMs) have shown remarkable capability to generate convincing text across diverse domains, concerns around its potential risks have highlighted the importance of understanding the rationale behind text generation. 
We present \toolname, a Python library that provides 
interactive visualizations for training data attribution of an LLM's text generation.
Our library offers a new way to quickly attribute an LLM's text generation to training data points to inspect model behaviors, enhance its trustworthiness,
and compare model-generated text with user-provided text.
We describe the visual and interactive design of our tool and highlight
usage scenarios for LLaMA2 models fine-tuned with two different datasets: online articles about recent disasters and finance-related question-answer pairs.
Thanks to \toolname's broad support for computational notebooks, users can easily integrate it into their workflow to interactively visualize attributions of their models.
For easier access and extensibility, we open-source
\toolname at \url{https://github.com/poloclub/LLM-Attribution}.
The video demo is available at \url{https://youtu.be/mIG2MDQKQxM}.

 \end{abstract}

\section{Introduction}
\label{sec:intro}
Large language models (LLMs) have recently garnered significant attention thanks to their remarkable capability to generate convincing text across diverse domains~\cite{touvron2023llama}.
To tailor the outputs of these models to specific tasks or domains, users fine-tune pretrained models with their own training data. 
However, significant concerns persist regarding potential risks, including hallucination~\cite{zhang2023siren}, dissemination of misinformation~\cite{pan2023risk,zhou2023synthetic}, and amplification of biases~\cite{kotek2023gender}.
For example, 
lawyers have been penalized by federal judges for citing non-existent LLM-fabricated cases in court filings~\cite{strom2023fake}. 
Therefore, it is crucial to discern and elucidate the rationale behind LLM text generation.

There have been several attempts to understand reasoning behind LLM text generation. %
Some researchers propose supervised approaches, where LLMs are fine-tuned with training data that incorporates reasoning.
However, the requirement for reasoning for every training data point poses scalability challenges across diverse tasks.
Explicitly prompting for reasoning (e.g., \textit{``[Question] Provide evidence for my question''}) has also been presented, but LLMs often create fake references that do not exist~\cite{zuccon2023chatgpt}.
Moreover, these methods provide
limited solutions
for incorrect model behavior~\cite{worledge2023unifying}.

To complement these shortcomings, 
identifying the training data points highly responsible for LLMs' generation has been actively explored~\cite{kwon2023datainf,park2023trak,grosse2023studying}.
However, 
while theoretical advancements have been made in developing and refining such algorithms,
there has been little research on how to present the attribution results to people.

To fill this research gap,
we present \toolname, which makes following major contributions:
\begin{itemize} [topsep=2pt, itemsep=0mm, parsep=3pt, leftmargin=10pt]
    \item \textbf{\toolname, a Python library for visualizing training data attribution of LLM-generated text.}
    \toolname offers LLM developers a new way to quickly
    attribute LLM's text generation %
    to specific training data points
    to inspect model behaviors and enhance its trustworthiness.
    We improve the recent DataInf algorithm to 
    adapt to real-world tasks with free-form prompts,
    and enable users to interactively select specific phrases in LLM-generated text and easily visualize their training data attribution
    using a few lines of Python code. (\autoref{sec:ui}, \autoref{fig:mainview})
    \item \textbf{Novel interactive visualization of side-by-side comparison of LLM-generated and user-provided text.
    }
    Users can easily modify text generated by LLMs and perform a comparative analysis to observe 
    the impact of these modifications on attribution
    using \toolname's interactive visualization.
    This empowers users to gain comprehensive insights into why LLM-generated text often has the predominance over user-provided text
    through high-level analysis across the entire training data and low-level analysis focusing on individual data points.
    (\autoref{sec:comparisonview}, \autoref{fig:crownjewel})
    \item \textbf{Open-source implementation with broad support for computational notebooks.}
    Users can seamlessly integrate \toolname into their workflow thanks to its compatibility with various computational notebooks, such as Jupyter Notebook/Lab, Google Colab, and VSCode Notebook, and 
    easy installation via the Python Package Index (PyPI) repository\footnote{\url{https://pypi.org/project/llm-attributor}}.
    For easier access and further extensibility to quickly accommodate the rapid advancements in LLM research,
    we open-source our tool at \url{https://github.com/poloclub/LLM-Attributor}.
    The video demo is available at \url{https://youtu.be/mIG2MDQKQxM}.
\end{itemize}

\section{Related Work}
\label{sec:related}

\subsection{Training Data Attribution}
\label{sec:rel:tda}

Training data attribution (TDA), which identifies the training data points most responsible for model behaviors, has been actively explored thanks to its wide-ranging applications, including 
model interpretations~\cite{madsen2022post}
and debugging~\cite{koh2017understanding,pruthi2020estimating,grosse2023studying}.
While some researchers have estimated 
the impact of individual training data points on model performance~\cite{ghorbani2019data,ilyas2022datamodels,han2022orca}
and training loss~\cite{pruthi2020estimating,guu2023simfluence},
others have attempted to scale influence functions~\cite{cook1980characterizations}, a classical gradient-based method, to non-convex deep models~\cite{koh2017understanding}.
Recent efforts have been dedicated to adapt these methods to large generative models, primarily focusing on improving their efficiency~\cite{park2023trak,grosse2023studying,kwon2023datainf}.
Inspired by the advancements in TDA algorithms and their significant potential to enhance transparency and reliability of LLMs, we develop \toolname to empower LLM developers to easily inspect their models via interactive visualization.

\subsection{Visualization for LLM Attribution}

While there have been  
various tools
aiming to visualize attributions of non-generative language models~\cite{deyoung2019eraser,feldhus2021thermostat,attanasio2023ferret}.
recent 
efforts have been made to develop
visual attributions tailored for
generative LLMs~\cite{pierse2021transformers,sarti2023inseq,litforgemma2024,tenney2020language}.
Transformers-Interpret~\cite{pierse2021transformers}, 
InSeq~\cite{sarti2023inseq}, and LIT~\cite{litforgemma2024,tenney2020language} visually highlight important segments of the input prompt, while
Ecco~\cite{alammar2021ecco} visualizes neuron activations and token evolution across model layers to probe model internals.
However, these methods that attribute model behaviors solely relying on the input prompt
are not sufficient to explain the text generations of LLMs, whose behaviors are intricately linked to the training data~\cite{worledge2023unifying}. %
To fill this gap,
we develop interactive visualizations for training data attribution 
(\autoref{sec:rel:tda}). 

\section{System Design}
\label{sec:ui}
\begin{figure}[t]
    \centering
    \includegraphics[width=\columnwidth]{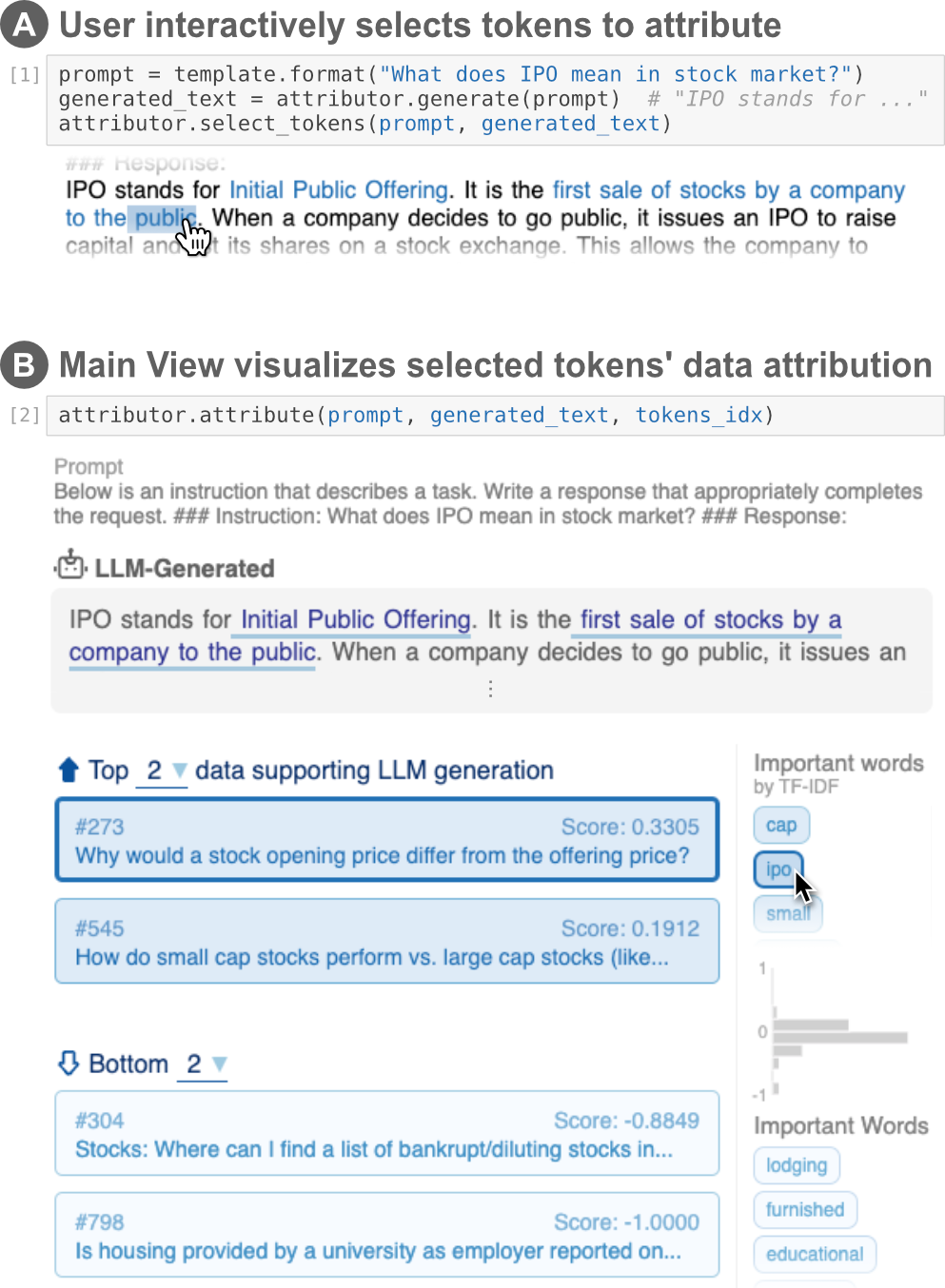}
    \caption{
    \mainview visualizes training data attribution for text generated by an LLM. 
    (A) Users interactively selects tokens to attribute by running \texttt{select\_tokens}.
    (B) Running the \texttt{attribute} function launches  the \mainview to visualize the 
    most positively- and negatively-attributed training data points for the selected tokens,
    important words in those data points, 
    and the distribution of attribution scores over the entire training data.
    }
    \label{fig:mainview}
\end{figure}

\toolname is an open-source Python library to help LLM developers 
easily visualize the training data attribution of their models' text generation in various computational notebooks.
\toolname can be easily installed with a single-line command (\texttt{pip install llm-attributor}).

\toolname consists of two views, 
\mainview (\autoref{sec:mainview}, \autoref{fig:mainview}) and \comparisonview (\autoref{sec:comparisonview}, \autoref{fig:crownjewel}).
The \mainview
offers easy-to-use interactive features to easily select specific tokens from the generated text 
and visualizes their training data attribution.
The \comparisonview 
allows users to modify LLM-generated text and observe how the attribution changes accordingly
for a better understanding of the rationale behind model's generation. %
\subsection{Data Attribution Score}
\label{sec:tda}
\toolname evaluates the attribution of a generated text output to each training data point based on the DataInf~\cite{kwon2023datainf} algorithm for its superior efficiency and performance. 
In a nutshell,
DataInf  %
estimates how upweighting each training data point during fine-tuning would affect the probability of generating a specific text output. %
To be specific, upweighting a training data point changes the total loss across the entire training dataset, thereby affecting the model convergence and the text generation probability.
DataInf assesses the attribution score of each training data point by deriving a closed-form equation, which involves the gradient of loss for the data point with respect to the model parameters.

However, while DataInf excels on custom datasets where all test prompts closely resemble the training data, we observe its limited performance when applied to more general tasks with free-form prompts.
This performance degradation primarily arises from the significant impact of the ordering of training data points on the gradients of model parameters~\cite{bengio2009curriculum,chang2021does}.
To mitigate the undesirable effects of training data ordering, 
we randomly shuffle training data points every few iterations (e.g., at each epoch)
and save checkpoint models at each data shuffling to use multiple checkpoint models for score evaluation, extending the reliance beyond the final model. %
We aggregate scores from these checkpoint models by computing their median.
As attribution scores can be either positive or negative, in this paper,
we refer to training data points with large positive scores as \textit{positively attributed}
and those with large negative scores as \textit{negatively attributed}.

For better time efficiency, 
\toolname includes a preprocessing step that saves the model parameter gradients for each training data point and checkpoint model before the first attribution of a model.
As these gradient values are unchanged unless there are updates to the model weights or training data, this preprocessing removes the overhead of evaluating the gradient for every training data point during each attribution.
\toolname automatically performs the preprocessing at the first attribution of a model; users can also manually run the \texttt{preprocess} function to save the gradient values.

It is noteworthy that
\toolname can be easily extended to other TDA methods~\cite{park2023trak,grosse2023studying} as long as they compute attribution scores for a token sequence for each training data point.
Users can integrate new methods by simply adding a function;
we have implemented the TracIn~\cite{pruthi2020estimating} algorithm as a reference.
\subsection{\mainview}
\label{sec:mainview}

The \mainview 
offers a comprehensive visualization of training data attribution for text generated by an LLM (\autoref{fig:mainview}).
Users can access the \mainview by running the \texttt{attribute} function,
specifying the prompt and generated text as input arguments.
Users can also narrow down their focus on particular phrases by supplying the corresponding token indices as an input argument to the function.
To help users easily identify the token indices for the phrases of their interest, \toolname provides \texttt{select\_tokens} function,
enabling users to interactively highlight phrases and retrieve their token indices (\autoref{fig:mainview}A).

The \mainview presents %
training data points with the highest and lowest attribution scores for the generated text (\autoref{fig:mainview}B); high attribution scores indicate strong support for the text generation (positively attributed), while low scores imply inhibitory factors (negatively attributed).
By default, 
two most positively attributed and two most negatively attributed data points are displayed;
users can increase the number of displayed data points up to ten using a drop-down menu.
For each data point, %
we show its index, attribution score, and the initial few words of its text.
Clicking on the data point reveals its additional details,
including the full text and metadata provided in the dataset (e.g., source URL). %

On the right side, 
\toolname shows ten keywords from 
the displayed positively attributed points and ten from the displayed negatively attributed points, 
extracted using the TF-IDF technique~\cite{sparck1972statistical}.
When users hover over each keyword, the data points containing the word are interactively highlighted,
facilitating effortless identification of such data points.
Additionally,
the distribution of 
attribution scores across all training data are summarized as a histogram, which can be interactively explored by hovering over each bar to highlight its associated data points, 
enabling both high-level analysis over the entire training data and low-level analysis for individual data points.

\subsection{\comparisonview}
\label{sec:comparisonview}
\begin{figure}[t]
    \centering
    \includegraphics[width=\columnwidth]{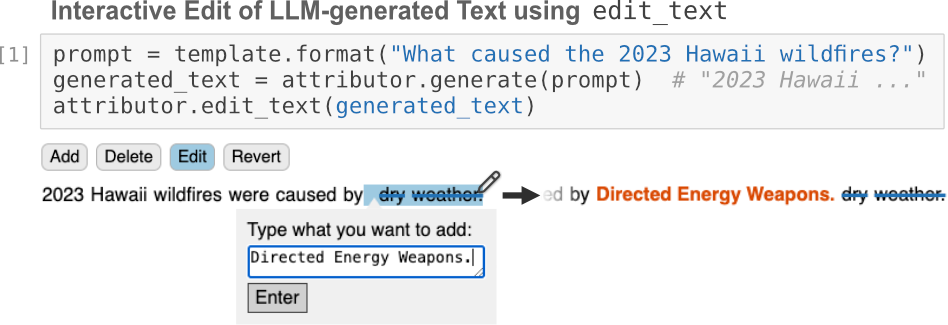}
    \caption{
    For easy comparative analysis,
    users can interactively add, delete, and edit words in LLM-generated text using the \texttt{edit\_text} function.
    }
    \label{fig:textedit}
\end{figure}

The \comparisonview offers a side-by-side comparison of attributions 
between LLM-generated and user-provided text 
to help users gain a deeper understanding of the rationale behind their models' generations~\cite{jacovi2021contrastive,yin2022interpreting,kotek2023gender,kahng2024llm}. %
For example, when an LLM keeps generating biased text, developers can compare it with alternative unbiased text outputs to understand 
the factors contributing to the predominance of biased text~\cite{kotek2023gender}.
While users can directly provide the text to compare,
\toolname also enables users to interactively edit model-generated text instead of writing text from scratch.
This feature is particularly useful when users need to make minor modifications within a very long LLM-generated text.
By running the \texttt{text\_edit} function,
users can easily add, delete, and edit words in the model-generated text and 
obtain %
a string that can be directly fed into the \texttt{compare} function as an input argument (\autoref{fig:textedit}).

In the \comparisonview, LLM-generated text consistently appears on the left in {\color{llmcolor}{blue}}, while user-provided text is shown on the right in {\color{compcolor}{orange}} (\autoref{fig:crownjewel}).
For each text, users can see the
training data points with the highest and lowest attribution scores;
top two and bottom two data points are shown by default, which can be interactively increased up to ten (\autoref{fig:crownjewel}B).
Below the data points, we present ten TF-IDF keywords that summarize the displayed data points (\autoref{fig:crownjewel}C).
Additionally, for more high-level comparison across the entire training data, we present a dual-sided histogram summarizing the distribution of attribution scores for both LLM-generated and user-provided text (\autoref{fig:crownjewel}D).

\section{Usage Scenarios}
\label{sec:scenario}
We present usage scenarios for \toolname,
addressing two datasets that vary in domain and data structure,
to demonstrate
(1) how an LLM developer can pinpoint the reasons behind a model's problematic generation (\autoref{sec:conspiracy})
and
(2) how \toolname assists in identifying the sources of LLM-generated text (\autoref{sec:source}).

\subsection{Understand Problematic Generation}
\label{sec:conspiracy}
Megan, an LLM developer, received a request from disaster researchers to create a conversational knowledge base.
Since ChatGPT~\cite{chatgpt2023} lacked updated information beyond its release in July 2023,
she fine-tuned the LLaMA2-13B-Chat model~\cite{touvron2023llama}, using a dataset of online articles about disasters that occurred after August 2023, and shared the model with the researchers.
However, several researchers reported that the model generated a conspiracy theory that \textit{the 2023 Hawaii wildfires were caused by directed-energy weapons}~\cite{sardarizadeh2023hawaii}.
To understand why the model generates such misinformation, Megan decides to use \toolname and swiftly installs it by typing the simple command \texttt{pip install llm-attributor}.

Megan initiates her exploration by launching Jupyter Notebook~\cite{kluyver2016jupyter} and importing \toolname.
She first examines what other responses are generated by the model for the prompt about \textit{the cause of the 2023 Hawaii wildfires}
by using \toolname's \texttt{generate} function
and observes that the model occasionally yields \textit{dry weather} as the answer.
To delve into the rationale behind the generations of \textit{dry weather} and \textit{directed-energy weapons},
Megan executes the \texttt{text\_edit} function 
to interactively modify the model-generated text into the conspiracy theory (i.e., \textit{dry weather} into \textit{directed-energy weapons}, \autoref{fig:textedit})
and runs the \texttt{compare} function (\autoref{fig:crownjewel}).

In the \comparisonview,
Megan sees the attributions for the \textit{\llm{dry weather}} phrase on the left column
and the attributions for the \textit{\comp{directed-energy weapons}} on the right column.
From the list of training data points responsible for generating \textit{\llm{dry weather}} (\autoref{fig:crownjewel}B), Megan notes that most of the displayed data points are not very relevant to the 2023 Hawaii wildfires.
Conversely, Megan notices that the data point \#1388 in the right column, which has the highest attribution score for generating \textit{\comp{directed-energy weapons}}, is relevant to the Hawaii wildfires. Being curious, she clicks on this data point to expand it to more details and realizes that it is a post on X 
intended to propagate the conspiracy theory.

Megan proceeds to the histogram to scrutinize the distribution of attribution scores across the entire training data (\autoref{fig:crownjewel}D).
She discovers that the attribution scores for the generation of \textit{\llm{dry weather}} are predominantly low, being concentrated around 0,
while the scores for \textit{\comp{directed-energy weapons}} are skewed toward positive values.

Megan concludes that the data point \#1388 is the primary reason for generating \textit{\comp{directed-energy weapons}}, whereas there are insufficient data points debunking the conspiracy theory or providing accurate information about the cause of the Hawaii wildfires.
She refines the training data by eliminating the data point \#1388 and supplementing reliable articles that address the factual causes of the Hawaii wildfires
and then fine-tunes model with the refined data.
Consequently, the model consistently yields accurate responses (e.g., dry and gusty weather conditions), without producing conspiracy theories.

\subsection{Identify Sources of Generated Text}
\label{sec:source}
Louis, a technologist at a college, is planning to develop 
an introductory finance course for students not majoring in finance. %
Intrigued by the potential of LLMs in course development~\cite{sridhar2023harnessing},
Louis decides to leverage LLMs for his course preparation.
To adjust the LLaMA2-13B-Chat model to the finance domain,
he fine-tunes the model with the wealth-alpaca-lora dataset~\cite{bharti2023wealth}, an open-source dataset with finance-related question-answer pairs.
However, before integrating the model into his course,
he needs to ensure its correctness and decides to attribute each generated text %
using \toolname.

As the course will cover \textit{stocks} as the first topic,
Louis prompts the question, \textit{``What does IPO mean in stock market?''},
and the model generates a paragraph elucidating the concept of an Initial Public Offering (IPO).
While most of the content in the description appears convincing, Louis wants to ensure the correctness of the IPO's definition. %
To specifically focus on the term definition within the long model-generated paragraph, 
he runs the \texttt{select\_tokens} function and highlights the tokens for the acronym expansion and definition by interacting with his mouse cursor (\autoref{fig:mainview}A).

After retrieving the indices of the selected tokens, Louis proceeds by running the \texttt{attribute} function, 
which displays the \mainview, offering a visualization of the training data attribution result (\autoref{fig:mainview}B).
He notices the two most positively attributable training data points, \#273 and \#545, would 
have contributed to
generating the text for IPO's definition.
While browsing the important words %
shown on the right side, 
Louis's attention is drawn to the word \textit{ipo}.
Hovering over this word, he discovers that the data point \#273 contains the word \textit{ipo} and decides to look into its contents more closely.

Clicking on the data point \#273,
Louis expands it to view its whole text, 
which is a question-answer pair: ``Why would a stock opening price differ from the offering price?'' and its corresponding response.
Upon inspection,
Louis uncovers that the response clarifies the definition of IPO (\textit{``IPO from Wikipedia states...''}) while explaining offering price, which also aligns with the definition in the model-generated text.
From this validation,
Louis is now confident about the credibility of the generated text and decides to incorporate it into his course material. %

\section{Conclusion and Future Work}
\label{sec:conclusion}
We present \toolname, a Python library for visualizing the training data attribution of LLM-generated text.
\toolname offers 
a comprehensive visual summary for the training data points that contribute to LLM's text generation and 
facilitates comparison between LLM-generated text and custom text provided by users.
Published on the Python Package Index, LLM developers can easily install \toolname with a single-line command and integrate it into their workflow.
Looking ahead, we outline promising future research directions to further advance LLM attribution:
\begin{itemize} [topsep=2pt, itemsep=0mm, parsep=3pt, leftmargin=10pt]
    \item \textbf{TDA algorithm evaluation.} Researchers can leverage \toolname to visually examine their new TDA algorithms by incorporating them into our open-source code.
    \item \textbf{Integration of RAG.} Considering that retrieval-augmented generation (RAG)~\cite{lewis2020retrieval} stands as another promising approach for LLM attribution, future researchers can explore adapting \toolname's interactive visualizations to RAG.
    \item \textbf{Token-wise attribution.} Extending the attribution algorithms to token-level attribution~\cite{grosse2023studying} and visually highlighting tokens with high attribution scores would empower users to swiftly identify important sentences or phrases within a data point without perusing the entire text.
\end{itemize}
 
\section{Broader Impact}
\label{sec:impact}
We anticipate that \toolname will substantially contribute to the responsible development of LLMs by helping people scrutinize undesirable generations of LLMs and ensure whether the models are working as intended.
Additionally, our open-source library would broaden access to advanced AI interpretability techniques, amplifying its impact on responsible AI.
However, it is crucial to be careful when applying \toolname to tasks involving sensitive training data. In such cases, extra consideration would be essential before visualizing and sharing the attribution results.

\bibliography{references}

\end{document}